\def\year{2022}\relax
\documentclass[letterpaper]{article} 
\usepackage{aaai22}  
\usepackage{times}  
\usepackage{helvet}  
\usepackage{courier}  
\usepackage[hyphens]{url}  
\usepackage{graphicx} 
\usepackage{natbib}  
\usepackage{caption} 
\DeclareCaptionStyle{ruled}{labelfont=normalfont,labelsep=colon,strut=off} 
\usepackage{algorithm}
\usepackage{algorithmic}
\usepackage{graphicx}
\usepackage[caption=false]{subfig}
\DeclareUnicodeCharacter{0441}{c}
\usepackage{newfloat}
\usepackage{listings}
\floatstyle{ruled}
\newfloat{listing}{tb}{lst}{}
\floatname{listing}{Listing}
\title{Interactive Machine Learning for Image Captioning}
\author {
    Mareike Hartmann,\textsuperscript{\rm 1} \thanks{Accepted at the AAAI-22 Workshop on Interactive Machine Learning (IML@AAAI’22)}
    Aliki Anagnostopoulou,\textsuperscript{\rm 1}
    Daniel Sonntag \textsuperscript{\rm 1,2}
}
\affiliations {
    \textsuperscript{\rm 1} German Research Center for Artificial Intelligence (DFKI), Germany\\
    \textsuperscript{\rm 2} Applied Artifical Intelligence (AAI), Oldenburg University, Germany\\
    mareike.hartmann@dfki.de, aliki.anagnostopoulou@dfki.de, daniel.sonntag@dfki.de
}

\usepackage{bibentry}

\begin{document}

\maketitle

\begin{abstract}
We propose an approach for interactive learning for an image captioning model. As human feedback is expensive and modern neural network based approaches often require large amounts of supervised data to be trained, we envision a system that exploits human feedback as good as possible by multiplying the feedback using data augmentation methods, and integrating the resulting training examples into the model in a smart way. This approach has three key components, for which we need to find suitable practical implementations: feedback collection, data augmentation, and model update. We outline our idea and review different possibilities to address these tasks.
\end{abstract}

\section{Introduction}
Our goal is to improve an image captioning system \cite{biswasBS20} by learning from human feedback. The envisioned use case is an image captioning system initially trained on publicly available image captioning data, e.g., the MS COCO dataset \cite{lin2014microsoft}, that can be adjusted to user-specific data based on corrective feedback provided by the end-user, ideally after deployment. To make use of user-specific feedback in an efficient way, we suggest to use data augmentation techniques to build additional training examples based on the user feedback, and then find the most suitable way to update model parameters based on the additional training data. This approach raises three research questions that we plan to address: 
\begin{enumerate}
    \item What type of feedback is most useful and how can it be collected?
    \item Which augmentation strategies are most useful to maximize the impact of the feedback?
    \item How can the feedback best be integrated into the training process?
\end{enumerate}
We expect the answers to these questions to be inter-dependent, as, e.g., some augmentation strategies might fit well with specific types of feedback. Also, best solutions might differ depending on the specific type of captioning model used. In the following, we briefly review two image captioning architectures that we plan to work with, and then discuss several ideas for implementing our approach. 

\begin{figure}[t!]
  \includegraphics[width=\linewidth]{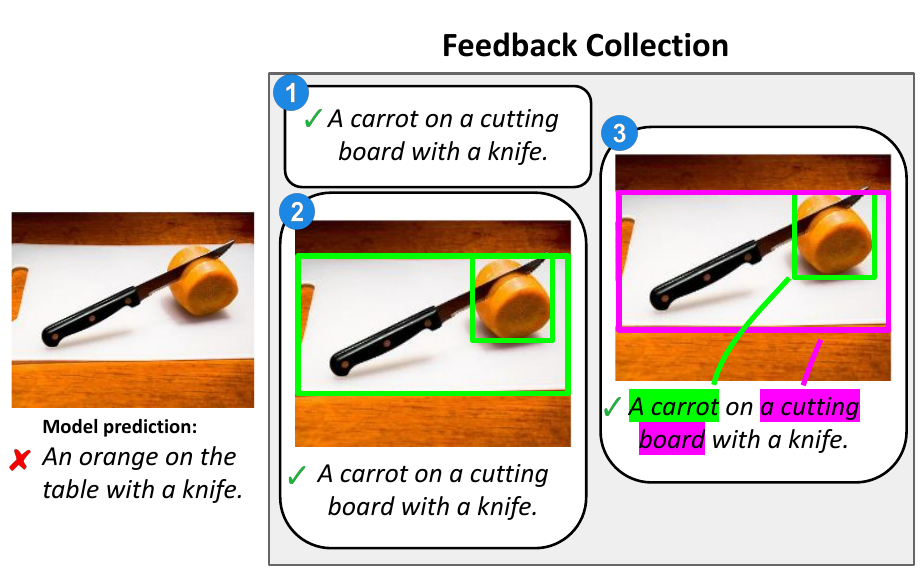}
  \caption{Examples for three different types of feedback: (1) corrected caption, (2) corrected caption with image annotations, (3) additional explicit alignment between objects and corrected caption words. }
  \label{fig:feedback}
\end{figure}

\section{Approach}
Our interactive learning approach for image captioning aims at improving a model based on user feedback, focusing on feedback collection, data augmentation, and feedback integration, rather than developing a new model architecture for image captioning. We consider using two different types of captioning models for our study: standard encoder-decoder models, such as Show-Attend-and-Tell \cite{xu2015show} or the Top-Down Bottom-Up Attention model \cite{anderson2018bottom} generate captions by feeding an attention-weighted representation of visual features, e.g., the output of a pre-trained object detection model, to the decoder at each time-step. The second family of models explicitly forces the decoder to generate words that are grounded in the image, which means that they refer to concepts detected in the image in a first step \cite{lu2018neural,cornia2019show,chen2020say,chen2021human}. The former models can be trained on pairs of images and corresponding captions. The latter models require finer-grained supervision, such as alignments between image regions and (sub)phrases, or scene graphs, but we expect them to be more suitable for our purpose, as they seem to offer a more direct way of integrating user feedback.

\begin{figure}[t]
  \includegraphics[width=\linewidth]{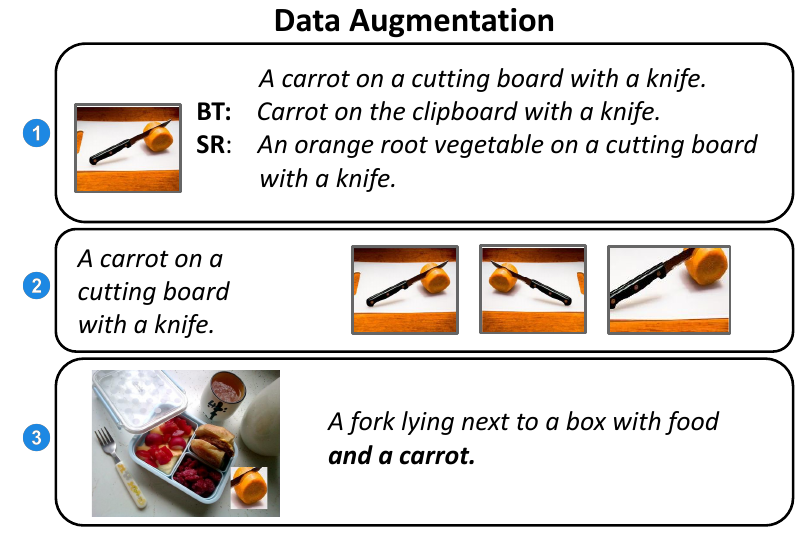}
  \caption{Examples of three different types of data augmentation: (1) caption-based including back-translation (BT) and synonym replacement (SR), (2) image-based, and (3) multi-modal by inserting patches cropped from a different image and adjusting the caption accordingly (added tokens are marked in bold).}
  \label{fig:data_augmentation}
\end{figure}

\paragraph{Feedback collection}
The first step is the collection of user feedback for model predictions on user-specific data. The user could provide different types of feedback of varying complexity (see Figure \ref{fig:feedback}), that might require more or less (cognitive) user efforts. Our goal should be to find a good balance between collecting rich feedback while not straining the user too much and keeping them engaged in the task, which makes it obligatory to implement a well-suited user interface. The least complex feedback type (in terms of user effort and requirements to the collection interface) would be a corrected image caption. Richer feedback could be collected by asking the user to additionally mark or select objects or image regions that the model incorrectly omitted from the generated caption, or even explicitly mark the alignment between corrected caption words and corresponding images. This type of feedback might be most useful for the controlled generation models. 

Another question to address is whether we can increase the impact of user feedback on model performance by requesting feedback for specific examples selected by a suitable deep active learning acquisition function  \cite{asghar2017deep,siddhant2018deep,lowell2019practical}. \citet{shen2019learning}
propose to collect feedback via an agent that learns to ask questions about specific visual concepts that it is uncertain about, which are answered by a human and used to construct new training examples. Their findings suggest that answering questions reduces human cost compared to providing a new caption and might be a reasonable alternative way of feedback collection to be explored.

\paragraph{Data augmentation}
An image-caption example paired with user feedback initially constitutes one additional training example for the captioning model. In the data augmentation step, we want to exploit the feedback to generate a larger amount of additional training examples. To this end, different augmentation strategies are applicable, either caption-based, image-based, or a combination of both modalities (see Figure \ref{fig:data_augmentation}). 

For \emph{image-based} augmentation, several image transformations such as cropping, warping, and flipping have previously been applied for image captioning \cite{wang2018image,takahashi2019data,katiyar2021image}. Such image transformations are possible, but might introduce noise in the form of mismatch between image and caption, for example if spatial relations are referred to in the caption (\emph{... is to the right of ...}). 

For \emph{caption-based} augmentation, \citet{atliha2020text} explored synonym replacement \cite{mccarthy-navigli-2007-semeval,character-level} and paraphrasing using a pre-trained language model \cite{devlin-etal-2019-bert}. Other text augmentation methods might be applicable, including random insertion/deletion/swapping of words \cite{EDA}, and backtranslation \cite{sennrich-etal-2016-improving}. For both modalities, retrieval-based augmentation from additional resources is possible as well \cite{li2021similar}. 

Another research direction is \emph{multi-modal} augmentation, where instead of modifying either caption or image, both modalities are modified at the same time. \citet{feng2021survey} bring up the idea of combining caption editing with image manipulation based on the CutMix method \cite{yun2019cutmix}, which augments data for image classification and object localization by inserting image patches (showing parts of objects) cut out from a different image, and respectively adjusting the label distribution. An adaptation for image captioning would require a re-write of the caption such that it correctly describes the modified image, which could either be done by replacing single words or by combining parts of different captions, similar to instance crossover augmentation proposed by  \citet{DBLP:journals/corr/abs-1909-11241}.

\paragraph{Model update}
Once we generated a reasonable amount of training data based on the user feedback, we need to update the model based on these new examples. Instead of re-training the model from scratch on the augmented training dataset \cite{shen2019learning}, we are more interested in the continual learning paradigm of batch-wise model updates, that allows to adjust the model to new information more efficiently (see \citet{ratt} for continual learning approaches applicable for image captioning). \citet{ling2017teaching} propose an approach for interactive image captioning that integrates phrase-level human feedback using reinforcement learning methods \cite{rennie2017self}, which we plan to compare to gradient-based methods. The main challenges in the continual learning setting include avoiding catastrophic forgetting of previously learned knowledge \cite{goodfellow2013empirical}, expanding the decoder to account for user-specific vocabulary, and integrating information about novel objects not observed previously \cite{hendricks2016deep}. For the latter problem, we plan to include experiments with inference-time strategies for novel object captioning \cite{anderson2017guided,zheng2019intention}. 

\section*{Acknowledgments}
The research was funded by the XAINES project (BMBF, 01IW20005) and the pAItient project (BMG, 2520DAT0P2).

\bibliography{aaai22}


\end{document}